\pdfoutput=1
\documentclass{rspublic}
\usepackage{graphicx,subfigure}

\theoremstyle{definition}
\newtheorem{defn}{Definition}
\newtheorem{prob}{Problem}

\begin{document}

\title[On Landmark Selection and Sampling]{On Landmark Selection and Sampling in High-Dimensional Data Analysis}

\author[M.-A. Belabbas and P. J. Wolfe]{Mohamed-Ali Belabbas and Patrick J. Wolfe\thanks{Author and address for correspondence: Statistics and Information Sciences Laboratory, Harvard University, Oxford Street, Cambridge, MA 02138, USA (wolfe@stat.harvard.edu)\vspace{2\baselineskip}}}

\affiliation{Statistics and Information Sciences Laboratory, Harvard University \\ Oxford Street, Cambridge, MA 02138, USA}

\label{firstpage}

\maketitle

\begin{abstract}{dimension reduction; kernel methods; low-rank approximation; machine learning; Nystr\"om extension}%
In recent years, the spectral analysis of appropriately defined kernel matrices has emerged as a principled way to extract the low-dimensional structure often prevalent in high-dimensional data.  Here we provide an introduction to spectral methods for linear and nonlinear dimension reduction, emphasizing ways to overcome the computational limitations currently faced by practitioners with massive datasets.  In particular, a data subsampling or landmark selection process is often employed to construct a kernel based on partial information, followed by an approximate spectral analysis termed the Nystr\"om extension.  We provide a quantitative framework to analyse this procedure, and use it to demonstrate algorithmic performance bounds on a range of practical approaches designed to optimize the landmark selection process.  We compare the practical implications of these bounds by way of real-world examples drawn from the field of computer vision, whereby low-dimensional manifold structure is shown to emerge from high-dimensional video data streams.
\end{abstract}

\section{Introduction}
\label{sec:intro}

In recent years, dramatic increases in available computational power and data storage capabilities have spurred a renewed interest in dimension reduction methods.  This trend is illustrated by the development over the past decade of several new algorithms designed to treat nonlinear structure in data, such as isomap (Tenenbaum {\em et al.}~2000), spectral clustering (Shi \&~Malik~2000), Laplacian eigenmaps (Belkin \&~Niyogi~2003), Hessian eigenmaps (Donoho \&~Grimes~2003) and diffusion maps (Coifman {\em et al.}~2005).  Despite their different origins, each of these algorithms requires computation of the principal eigenvectors and eigenvalues of a positive semi-definite kernel matrix.

In fact, spectral methods and their brethren have long held a central place in statistical data analysis. The spectral decomposition of a positive semi-definite kernel matrix underlies a variety of classical approaches such as principal components analysis, in which a low-dimensional subspace that explains most of the variance in the data is sought, Fisher discriminant analysis, which aims to determine a separating hyperplane for data classification, and multidimensional scaling, used to realize metric embeddings of the data.

As a result of their reliance on the exact eigendecomposition of an appropriate kernel matrix, the computational complexity of these methods scales in turn as the cube of either the dataset \emph{dimensionality} or \emph{cardinality} (Belabbas \&~Wolfe~2009).  Accordingly, if we write $\mathcal{O}(n^3)$ for the requisite complexity of an exact eigendecomposition, large and/or high-dimensional datasets can pose severe computational problems for both classical and modern methods alike.  One alternative is to construct a kernel based on partial information; that is, to analyse directly a set of `landmark' dimensions or examples that have been selected from the dataset as a kind of summary statistic.  Landmark selection thus reduces the overall computational burden by enabling practitioners to apply the aforementioned algorithms directly to a subset of their original data---one consisting solely of the chosen landmarks---and subsequently to extrapolate their results at a computational cost of $\mathcal{O}(n^2)$.

While practitioners often select landmarks simply by sampling from their data uniformly at random, we show in this article how one may improve upon this approach in a data-adaptive manner, at only a slightly higher computational cost.  We begin with a review of linear and nonlinear dimension-reduction methods in~\S\ref{sec:dimRed}, and formally introduce the optimal landmark selection problem in~\S\ref{sec:landSel}.  We then provide an analysis framework for landmark selection in~\S\ref{sec:analysisFramework}, which in turn yields a clear set of trade-offs between computational complexity and quality of approximation.  Finally, we conclude in~\S\ref{sec:caseStudy} with a case study demonstrating applications to the field of computer vision.

\section{Linear and nonlinear dimension reduction}
\label{sec:dimRed}

\subsection{Linear case: principal components analysis}

Dimension reduction has been an important part of the statistical landscape since the
inception of the field.  Indeed, though principal components analysis (PCA) was introduced more than a century ago, it still enjoys wide use among
practitioners as a canonical method of data analysis.  In recent years, however, the lessening costs of both computation and data storage have begun to alter the research landscape in the area of dimension
reduction:  massive datasets have gone from being rare cases to everyday burdens, with nonlinear
relationships amongst entries becoming ever more common.

Faced with this new landscape, computational considerations have become a necessary part of statisticians' thinking, and new approaches and methods are required to treat the unique problems posed by modern datasets.  Let us start by introducing some notation and explaining the principal(!) issues by way of a  simple
illustrative example.  Assume we are given a collection of $N$ data samples, denoted by the set $\mathcal{X}=\lbrace x_1,\ldots,x_N\rbrace$, with each sample $x_i$ comprising $n$ measurements. For example,
the samples $x_i$ could contain hourly measurements of  the temperature, humidity level and wind speed at a particular location over a period of a day; in this case $\mathcal{X}$ would
contain $24$ three-dimensional vectors.

\begin{figure}[t]
\subfigure[An example set of centred measurements, with projections on to each coordinate axis also shown]{
\includegraphics[width=0.5\columnwidth]{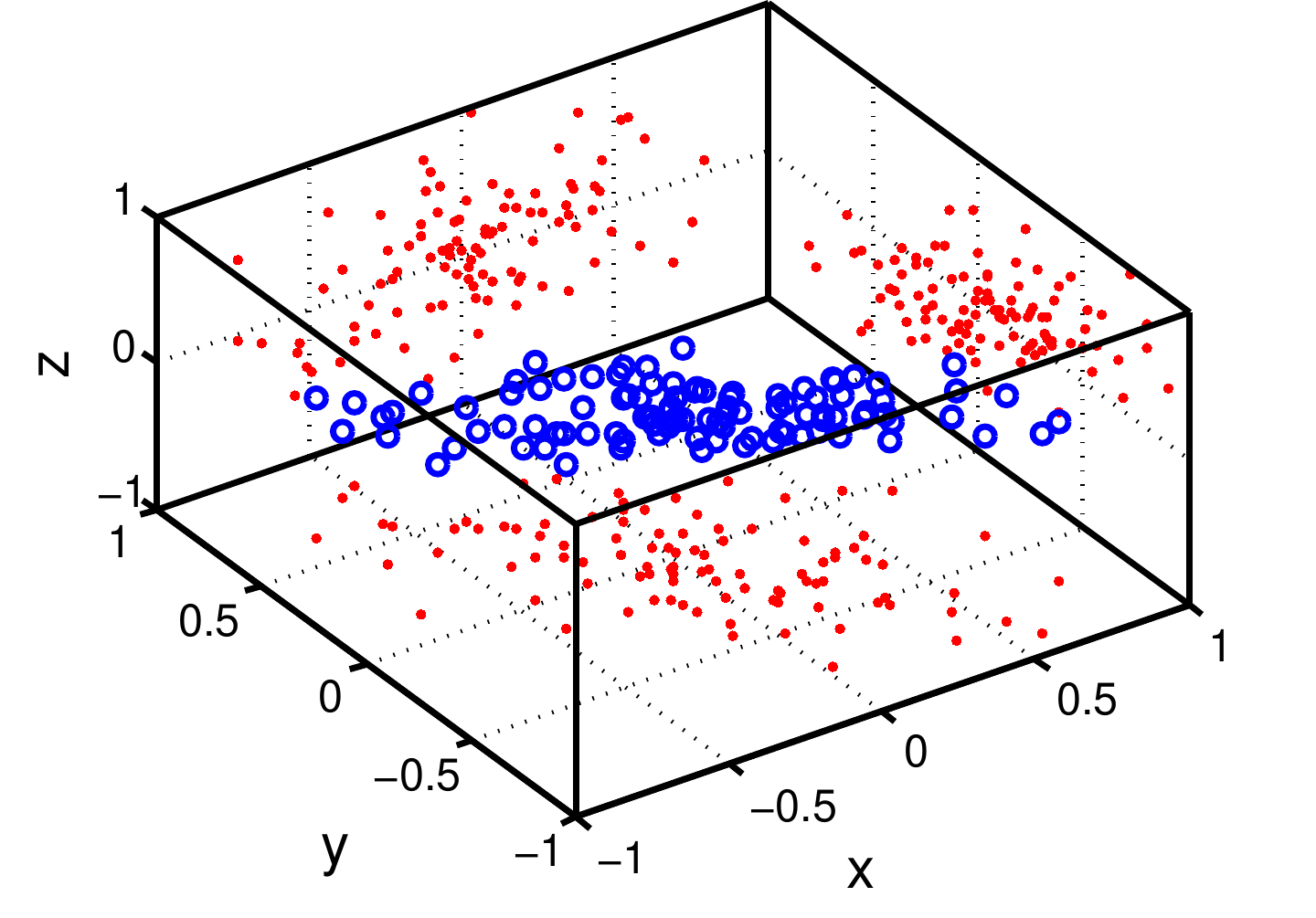}
\label{fig:PCAExample:subfiga}
}\hfill
\subfigure[PCA yields a plane indicating the directions of greatest variability of the data]{
\includegraphics[width=0.45\columnwidth]{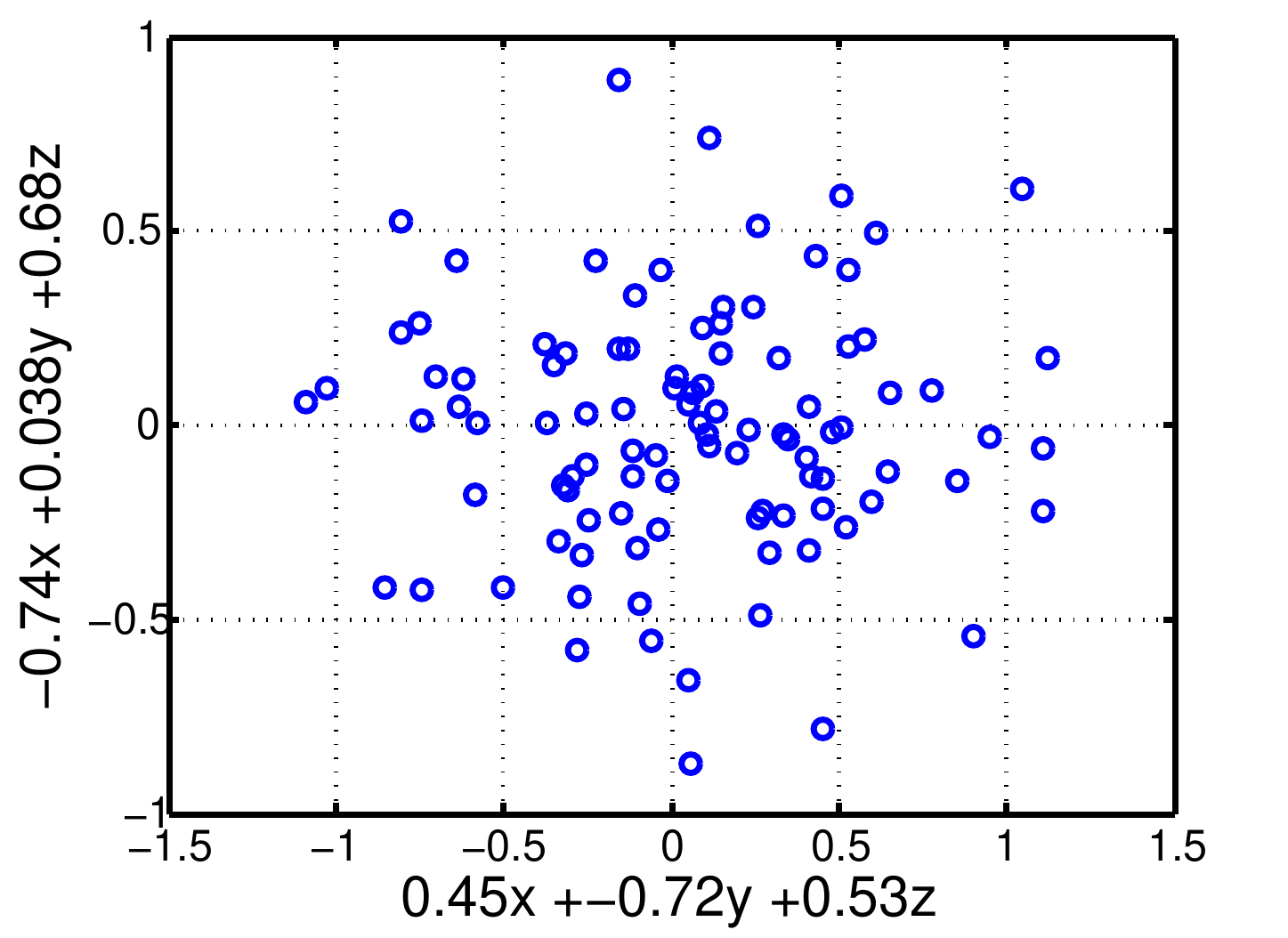}
\label{fig:PCAExample:subfigb}
}
\caption{\label{fig:PCAExample}Principal components analysis, with measurements in panel~\subref{fig:PCAExample:subfiga} expressed in panel~\subref{fig:PCAExample:subfigb} in terms of the two-dimensional subspace that best explains their variability}
\end{figure}

The objective of principal components analysis is to reduce the dimension of a given dataset by exploiting \emph{linear correlations} amongst its entries.  Intuitively, it is not hard to imagine that, say, as the temperature increases, wind speed might decrease---and thus retaining only the humidity levels and a linear combination
of the temperature and wind speed would be, up to a small error, as informative as knowing
all three values exactly.  By way of an example, consider gathering centred measurements (i.e., with the mean subtracted)
into a matrix $X$, with one measurement per column; for the example above, $X$ is of dimension $3 \times 24$. The method of principal components then consists of analysing the positive semi-definite kernel $Q= XX^T$ of outer products between all samples $x_i$ by way of its eigendecomposition $Q = U\Lambda U^T$, where $U:U^TU = I$ is an orthogonal matrix whose columns comprise the eigenvectors of $Q$, and $\Lambda$ is a diagonal matrix containing its real, nonnegative eigenvalues.  The eigenvectors associated with the largest eigenvalues of $Q$ yield a new set of variables according to $Y=U^TX$, which in turn provide the (linear) directions of greatest variability of the data (see figure~\ref{fig:PCAExample}).

\subsection{Nonlinear case: diffusion maps and Laplacian eigenmaps}
\label{sec:nonlin}

\begin{figure}[t]
\centering
\subfigure[`Fishbowl' data (sphere with top cap removed)]{
\includegraphics[width=0.5\columnwidth]{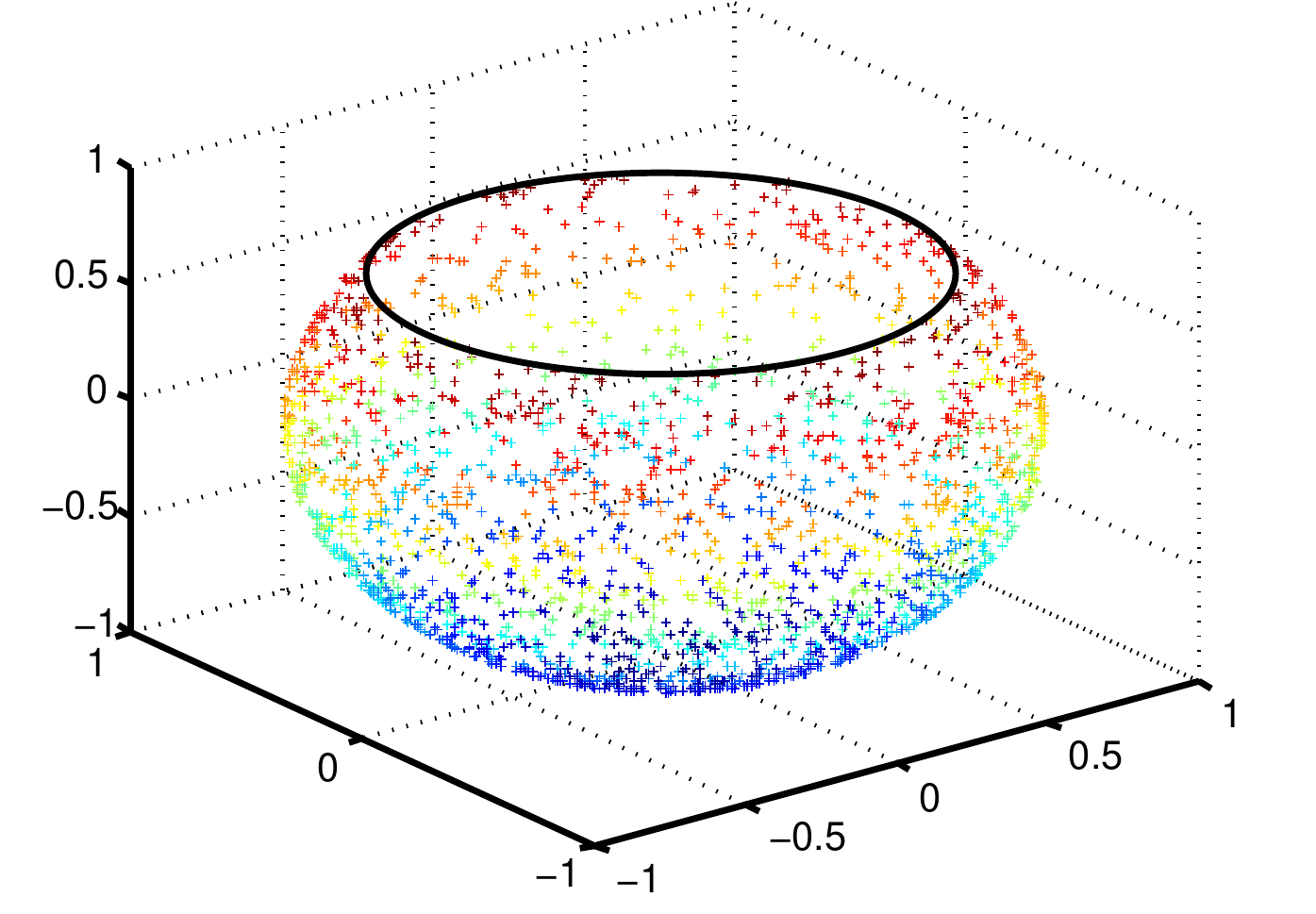}
\label{fig:fishbowlExample:subfig1}
}\\
\subfigure[PCA]{
\includegraphics[width=0.25\columnwidth]{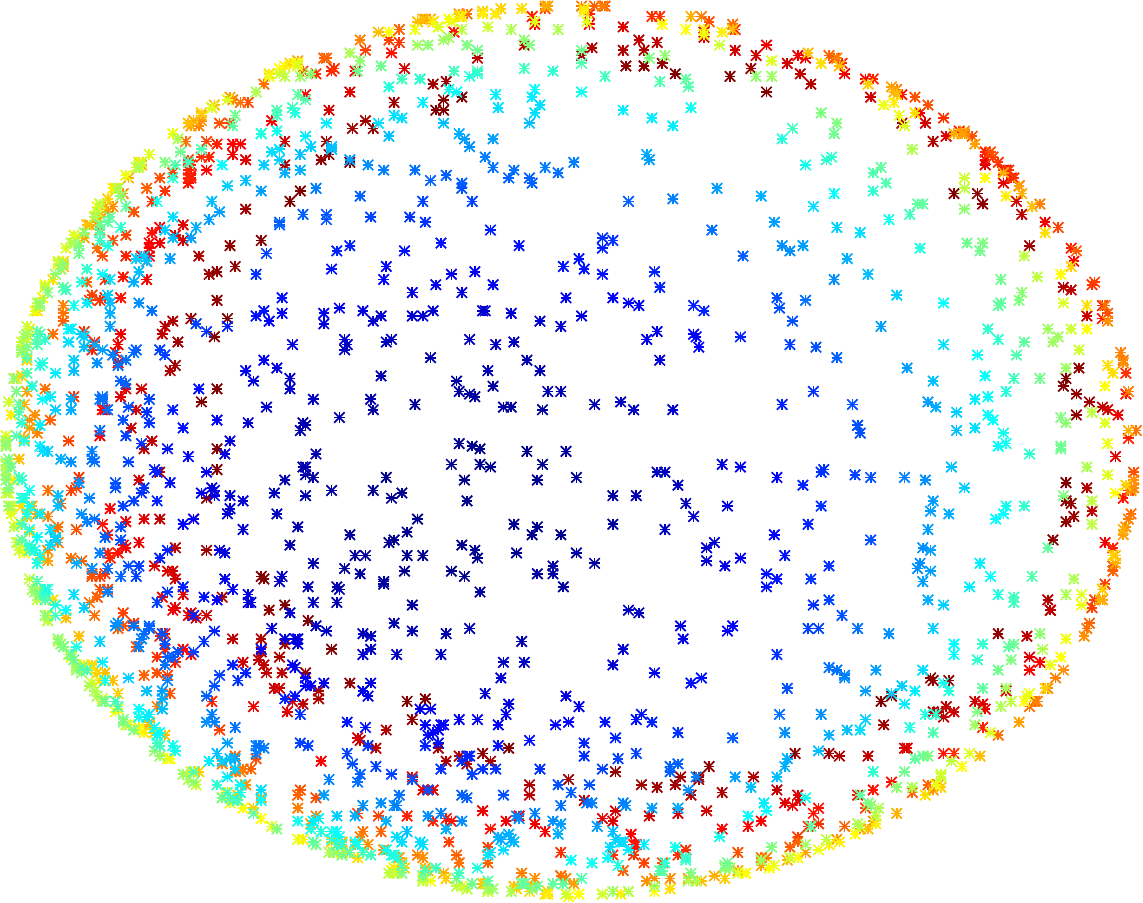}
\label{fig:fishbowlExample:subfig2}
}\hfill
\subfigure[Diffusion maps]{
\includegraphics[width=0.25\columnwidth]{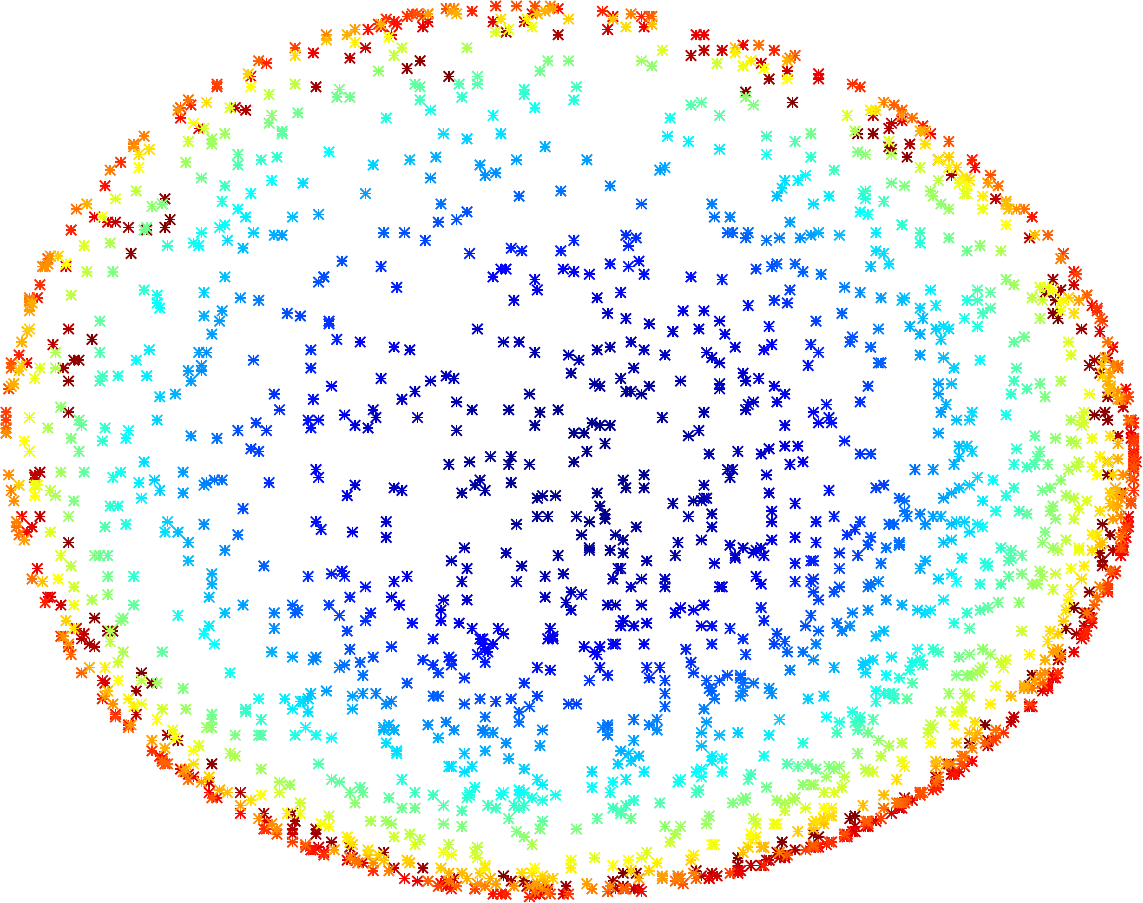}
\label{fig:fishbowlExample:subfig3}
}\hfill
\subfigure[Laplacian eigenmaps]{
\includegraphics[width=0.25\columnwidth]{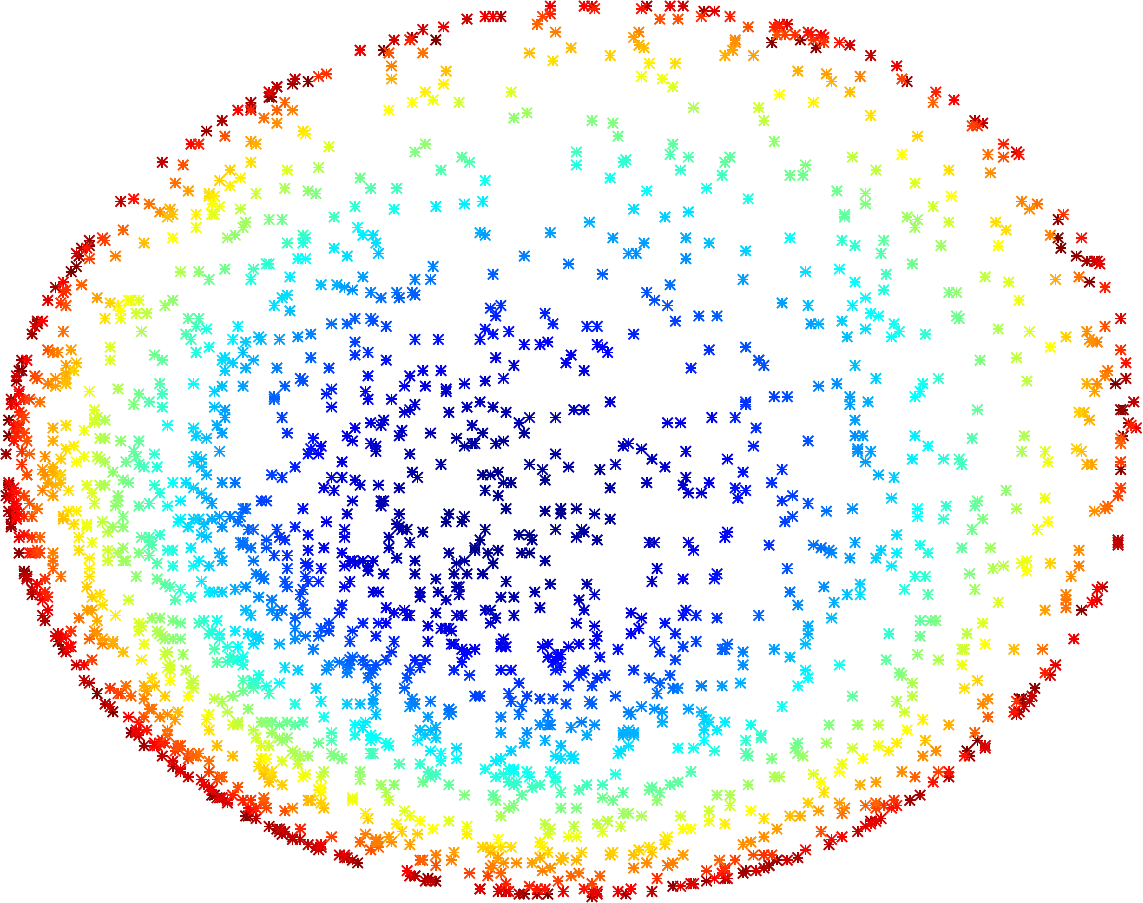}
\label{fig:fishbowlExample:subfig4}
}
\caption{\label{fig:fishbowlExample}Nonlinear dimension reduction, with contrasting embeddings of the data of panel~\subref{fig:fishbowlExample:subfig1} shown.  The two-dimensional linear embedding via PCA, shown in panel~\subref{fig:fishbowlExample:subfig2}, yields an overlap of points of different colour, indicating a failure to recover the nonlinear structure of the data.  Panels~\subref{fig:fishbowlExample:subfig3} and~\subref{fig:fishbowlExample:subfig4} show respectively the embeddings obtained by diffusion maps and Laplacian eigenmaps; each of these methods successfully recovers the nonlinear structure of the original dataset, correctly `unfolding' it in two dimensions.}
\end{figure}

In the above example, PCA will be successful if the relationship between wind speed
and temperature (for example) is linear. \emph{Nonlinear} dimension reduction refers to the case in which the relationships between variables are not linear, whereupon the method of principal components will fail to explain adequately any nonlinear co-variability present in the measurements.  An example dataset of this type is shown in figure~\ref{fig:fishbowlExample:subfig1}, consisting of points sampled from a two-dimensional disc stretched into a three-dimensional shape taking the form of a fishbowl.

In the same vein as PCA, however, most contemporary methods for nonlinear dimension reduction are based on the analysis of an appropriately defined positive semi-definite kernel.  Here we limit ourselves to describing two closely related methods that serve to illustrate the case in point: diffusion maps (Coifman {\em et al.}~2005) and Laplacian eigenmaps (Belkin \&~Niyogi~2003).

\subsubsection{Diffusion maps}
\label{sec:diffmaps}

Given input data $\mathcal{X}$ having cardinality $N$ and dimension $n$, along with parameters $\sigma > 0$ and $m$ a positive integer, the diffusion maps algorithm involves first forming a positive semi-definite kernel $Q$ whose $(i,j)$th entry is given by
\begin{equation}
\label{eq:defW}
Q_{ij}=e^{-\|x_i-x_j\|^2/2\sigma^2}\text{,}
\end{equation}
with $\|x_i-x_j\|$ the standard Euclidean norm on $\mathbb{R}^n$.  If we define a diagonal matrix $D$ whose entries are the corresponding row/column sums of $Q$ as $D_{ii}=\sum_{j}Q_{ij}$, the Markov transition matrix $P=D^{-1}Q$ is then computed.  This transition matrix describes the
evolution of a discrete-time diffusion process on the points of $\mathcal{X}$, where the transition
probabilities are given by~\eqref{eq:defW}, with multiplication of $Q$ by $D^{-1}$ serving
to normalize them.

As is well known, the corresponding transition matrix after $m$ time steps is simply given by the $m$-fold product of $P$ with itself; if we write $P^m = U \Lambda^m U^{-1}$, the principal eigenvectors and eigenvalues of this transition matrix are used to embed the data according to $Y=U \Lambda^m$.  However, note that, since $P$ is a stochastic matrix, its principal eigenvector is $[1 \,\,\, 1 \,\,\, \cdots \,\,\, 1]^T$, with corresponding eigenvalue equal to unity. This eigenvector-eigenvalue pair is hence ignored for purposes of the embedding, as it does not depend on $\mathcal{X}$.

Although $P=D^{-1}Q$ is not symmetric, its eigenvectors can equivalently be obtained via spectral analysis of the positive semi-definite kernel $\widetilde{Q} = D^{1/2}PD^{-1/2} = D^{-1/2}QD^{-1/2}$: if $(\lambda, u)$ satisfy $Pu =\lambda u$, then, if $\tilde u= D^{1/2}u$, we obtain
\begin{equation*}
  D^{1/2}Pu =\lambda D^{1/2}u \quad \Rightarrow \quad
  D^{1/2}PD^{-1/2}\,D^{1/2}u =\lambda \,D^{1/2}u \quad \Rightarrow \quad
  \widetilde{Q} \tilde u = \lambda \tilde u \text{.}
\end{equation*}
Hence from this analysis we see that $P$ and $\widetilde{Q}$ share identical eigenvalues, as well as eigenvectors related by a diagonal transformation.

\subsubsection{Laplacian eigenmaps}

Rather than necessarily computing a dense kernel $Q$ as in the case of diffusion maps, the Laplacian eigenmaps algorithm commences with the computation of a $k$-neighbourhood for each data point $x_i$; i.e., the $k$ nearest data points to each $x_i$ are found.  A \emph{weighted graph} whose vertices are the data points $\{x_1,x_2,\ldots,x_N\}$ is then computed, with an edge present between vertices $x_i$ and $x_j$ if and only if $x_i$ is among the $k$ closest points to $x_j$, or vice-versa.  The weight of each kernel entry is given by $Q_{ij}=e^{-\|x_i-x_j\|^2/2\sigma^2}$ if an edge is present in the corresponding graph, and $Q_{ij} = 0$ otherwise, and thus we immediately arrive at a \emph{sparsified} version of the diffusion maps kernel.

The embedding $Y$ is chosen to minimize the weighted sum of pairwise distances
\begin{equation}
\label{eq:LapEigMin}
\sum_{ij} \|y_i-y_j\|^2 Q_{ij} \text{,}
\end{equation}
subject to the normalization constraints $\| D^{1/2} y_i \| = 1$, where, as in the case of diffusion maps, $D$ is a diagonal matrix with
entries $D_{ii}=\sum_j Q_{ij}$.

Now consider the so-called combinatorial Laplacian of the graph, defined as the positive semi-definite kernel $L = D - Q$.  A simple calculation shows that the constrained minimization of~\eqref{eq:LapEigMin} may be reformulated as
\begin{equation*}
\operatornamewithlimits{argmin}_{Y^T D Y = I} \, \operatorname{tr}(Y^TLY) \text{,}
\end{equation*}
whose solution in turn will consist of the eigenvectors of $D^{-1}L$ with \emph{smallest} eigenvalues---from which we exclude, as in the case of diffusion maps, the solution proportional to $[1 \,\,\, 1 \,\,\, \cdots \,\,\, 1]^T$.  By the same argument as employed in~\S\ref{sec:dimRed}$\,\ref{sec:nonlin}\,$(\ref{sec:diffmaps}) above, this analysis is easily related to that of the \emph{normalized} Laplacian $D^{-1/2}LD^{-1/2}$.

\subsection{Computational considerations}

Recall our earlier assumption of a collection of $N$ data samples, denoted by the set $\mathcal{X}=\lbrace x_1,\ldots,x_N\rbrace$, with each sample $x_i$ comprising $n$ measurements.  An important point of the above analyses is that, in each case, the size of the kernel is dictated by
either the \emph{number} of data samples (diffusion maps, Laplacian eigenmaps) or their \emph{dimension} (PCA). Indeed, classical and modern spectral methods rely on either of the following:
\begin{description}
\item[Outer characteristics of the point cloud.] Methods such
  as PCA or Fisher discriminant analysis require the analysis of a kernel of dimension $n$, the
  extrinsic dimension of the data;
\item[Inner characteristics of the point cloud.] Multidimensional scaling and recent extensions that perform nonlinear embeddings of data points require the
  spectral analysis of a kernel of dimension $N$, the cardinality of
  the point cloud.
\end{description}

 In both sets of scenarios, the analysis of large kernels quickly induces a computational burden that is impossible to overcome with exact spectral methods, thereby motivating the introduction of landmark selection and sampling methods.

\section{Landmark selection and the Nystr\"om method}
\label{sec:landSel}

Since their introduction, and furthermore as datasets continue to increase in size and dimension, so-called landmark methods have seen wide use by practitioners across various fields.  These methods exploit the high level of redundancy often present in high-dimensional datasets by seeking a small (in relative terms) number of important examples or coordinates that summarize the most relevant information in the data; this amounts in effect to an adaptive compression scheme.  Separate from this subset selection problem is the actual solution of the corresponding spectral analysis task---and this in turn is accomplished via the so-called Nystr\"om extension (Williams \&~Seeger~2001; Platt~2005).

While the Nystr\"om reconstruction admits the unique property of providing, conditioned upon a set of selected landmarks, the minimal kernel completion with respect to the partial ordering of positive semi-definiteness, the literature is currently open on the question of optimal landmark selection.  Choosing the most appropriate set of landmarks for a specific dataset is a fundamental task if spectral methods are to successfully `scale up' to the order of the large datasets already seen in contemporary applications, and expected to grow in the future.  Improvements will in turn translate directly to either a more efficient compression of the input (i.e., fewer landmarks will be needed) or a more accurate approximation for a given compression size.  While choosing landmarks in a data-adaptive way can clearly offer improvement over approaches such as selecting them uniformly at random (Drineas \&~Mahoney~2005; Belabbas \&~Wolfe~2009), this latter approach remains by far the most popular with practitioners (Smola \&~Sch{\"o}lkopf~2000; Fowlkes {\em et al.}~2001, 2004; Talwalkar {\em et al.}~2008).

While it is clear that data-dependent landmark selection methods offer the potential of at least some improvement over non-adaptive methods such as uniform sampling (Liu {\em et al.}~2006), bounds on performance as a function of computation have not been rigorously addressed in the literature to date.  One important reason for this has been the lack of a unifying framework to understand the problems of landmark selection and sampling, and to provide approximation bounds and quantitative performance guarantees.  In this section we describe an analysis framework for landmark selection that places previous approaches in context, and show how it leads to quantitative performance bounds on Nystr\"om kernel approximation.

\subsection{Spectral methods and kernel approximation}

As noted earlier, spectral methods rely on low-rank approximations of appropriately defined positive semi-definite kernels.  To this end, let $Q$ be a real, symmetric kernel matrix of dimension $n$;
we write $Q \succeq 0$ to denote that $Q$ is positive semi-definite.  Any such kernel $Q \succeq 0$ can in turn be expressed in spectral coordinates as $Q=U \Lambda U^T$, where $U$ is an orthogonal matrix and $\Lambda=\operatorname{diag}(\lambda_1,\ldots,\lambda_n)$ contains the real, nonnegative eigenvalues of $Q$, assumed sorted in non-increasing order.

To measure the error in approximating a kernel $Q \succeq 0$, we require the following notion of unitary invariance (see, e.g.,~Horn \&~Johnson~(1990)).
\begin{defn}[Unitary Invariance]
A matrix norm $\| \cdot \|$ is termed \emph{unitarily invariant} if, for all matrices $U,V: U^TU = I, V^TV = I$, we have $\| U M V^T \| = \| M \|$ for every (real) matrix $M$.
\end{defn}
A unitarily invariant norm therefore depends only on the singular values of its argument, and for any such norm the optimal rank-$k$ approximation to $Q \succeq 0$ is given by $Q_k := U \Lambda_k U^T$ where $\Lambda_k = \operatorname{diag}(\lambda_1, \lambda_2, \ldots, \lambda_k,0,\ldots,0)$.  When a given kernel $Q$ is expressed in spectral coordinates, evaluating the quality of any low-rank approximation $\widetilde{Q}$ is a trivial task, requiring only an ordering of the eigenvalues.  As described in~\S\ref{sec:intro}, however, the cost of obtaining these spectral coordinates exactly is $\mathcal{O}(n^3)$, which is often too costly to be computed in practice.

To this end, methods that rely on either the \emph{extrinsic} dimension of a point cloud, or on the \emph{intrinsic} dimension of a set of training examples via its cardinality, impose a large computational burden.  To illustrate, let $x_1, x_2, \ldots, x_N \in \mathbb{R}^n$ comprise the data of interest.  `Outer' methods of the former category employ a rank-$k$ approximation of the matrix $Q := \sum_{i=1}^N x_ix_i^T$, which is of dimension $n$.  Alternatively, `inner' methods introduce an additional positive-definite function $q(x_i,x_j)$, such as $\langle x_i,x_j \rangle $ or $\exp(-\|x_i-x_j\|^2/2\sigma^2)$, and obtain a $k$-dimensional embedding of the data via the $N$-dimensional affinity matrix $Q_{ij} := q(x_i,x_j)$.

\subsection{The Nystr\"om method and landmark selection}

The Nystr\"om method has found many applications in modern machine learning and data analysis applications as a means of obtaining an approximate spectral analysis of the kernel of interest $Q$.  In brief, the method solves a matrix completion problem in a way that preserves positive semi-definiteness as follows.
\begin{defn}[Nystr\"om Extension]\label{defn:Nyst}
   Fix a subset $J \subset \lbrace 1,2,\ldots n \rbrace$ of cardinality $k < n$, and let $Q_{J}$ denote the corresponding principal  submatrix of an $n$-dimensional kernel $Q \succeq 0$. Take $J=\{ 1,2,\ldots ,k \}$ without loss of generality and partition $Q$ as follows:
  \begin{equation}
    \label{eq:partQ}
    Q=\left[\begin{array}{cc} Q_J & Y \\Y^T &
        Z\end{array}\right] \text{.}
  \end{equation}
The \emph{Nystr\"om extension} then approximates $Q$ by
\begin{equation}
\label{eq:partQtilde}
  \widetilde Q = \left[ \begin{array}{cc} Q_J & Y \\ Y^T & Y^T Q_J^{-1} Y \end{array}\right] \succeq 0 \text{.}
\end{equation}
Here $Q_J \in \mathbb{R}^{k \times k}$ and $Z \in \mathbb{R}^{(n-k) \times (n-k)}$ are always positive semi-definite, being principal submatrices of $Q\succeq 0$, and $Y$ is a rectangular submatrix of dimension $k \times (n-k)$.
\end{defn}
If we decompose $Q_J$ as $Q_J=U_J \Lambda_J U_J^T$, this corresponds to approximating the eigenvectors and eigenvalues of $Q$ by
\begin{equation*}
  \widetilde \Lambda = \Lambda_J, \quad \widetilde U = \left[ \begin{array}{c} U_J \\Y^T U_J \Lambda_J^{-1} \end{array}\right] \text{.}
\end{equation*}
We have that $\operatorname{rank}(\widetilde Q) \leq k$, and (noting that typically $k \ll n$) the complexity of reconstruction is of order $\mathcal{O}(n^2k)$. Approximate eigenvectors $\tilde U$ can be obtained in $\mathcal{O}(nk^2)$, and can be orthogonalized by an additional projection.

The Nystr\"om method thus serves as a means of completing a partial kernel, conditioned upon a selected subset $J$ of rows and columns of $Q$.  The landmark selection problem becomes that of choosing the subset $J$ of fixed
cardinality $k$ such that $\|Q - \widetilde Q \|$ is minimized for some unitarily invariant norm, with a lower bound given by $\|Q - Q_k\|$, where $Q_k$ is the optimal rank-$k$ approximation obtained by setting the $n-k$ smallest eigenvalues of $Q$ to zero.

According to the difference between~\eqref{eq:partQ} and~\eqref{eq:partQtilde}, the approximation error $\| Q-\widetilde Q\|$ can in general be expressed in terms of the Schur complement of $Q_J$ in $Q$, defined as $Z - Y^T Q_J^{-1}Y$ according to the conformal partition of $Q$ in~\eqref{eq:partQ}, and correspondingly for an appropriate permutation of rows and columns in the general case.

With reference to definition~\ref{defn:Nyst}, we thus have the optimal landmark selection problem as follows.
\begin{prob}[Optimal Landmark Selection]\label{prob:kernel}
  Choose $J$, with cardinality $|J|=k$, such that $\|Q - \widetilde Q \| = \|Z-Y^TQ_J^{-1}Y\|$ is minimized.
\end{prob}
It remains an open question as to whether or not, for any unitarily invariant norm, this subset selection problem can be solved in fewer than $\mathcal{O}(n^3)$ operations, the threshold above which the exact spectral decomposition becomes the best option. In fact, there is no known exact algorithm other than $\mathcal{O}(n^k)$ brute-force enumeration in the general case.

\section{Analysis framework for landmark selection}
\label{sec:analysisFramework}

Attempts to solve the landmark selection problem can be divided into two categories: deterministic methods that typically minimize some objective function in an iterative or stepwise greedy fashion (Smola \&~Sch{\"o}lkopf~2000; Ouimet \&~Bengio~2005; Liu {\em et al.}~2006; Zhang \&~Kwok~2009), for which the resultant quality of kernel approximation cannot typically be guaranteed, and randomized algorithms that instead proceed by sampling (Williams \&~Seeger~2001; Fowlkes {\em et al.}~2004; Drineas \&~Mahoney~2005; Belabbas \&~Wolfe~2009).  As we show in this section, those sampling-based methods for which relative error bounds currently exist can all be subsumed within a generalized stochastic framework that we term annealed determinant sampling.

\subsection{Nystr\"om error characterization}
It is instructive first to consider problem~\ref{prob:kernel} in more detail, in order that we may better characterize properties of the Nystr\"om approximation error.  To this end, we adopt the \emph{trace norm} $\| \cdot \|_{\operatorname{tr}}$ as our unitarily invariant norm of interest.
\begin{defn}[Trace Norm]
Fix an arbitrary matrix $M\in\mathbb{R}^{m \times n}$ and let $\sigma_i(M)$ denote its $i$th singular value.  Then the trace norm of $M$ is defined as
\begin{align}\notag
\| M \|_{\operatorname{tr}} & = \operatorname{tr}( \sqrt{M^T\!M}) = \sum_{i=1}^{\min(m,n)} \sigma_i(M)
\\ & \equiv \operatorname{tr}(Q) \text{ for $Q \succeq 0$} \label{eq:TraceNormEquivalence}
\text{.}
\end{align}
\end{defn}
Since any positive semi-definite kernel $Q  \succeq 0$ admits the Gram decomposition $Q = X^T \! X$, this implies the following relationship in Frobenius norm $\| \cdot \|_F$, to be revisited shortly:
\begin{equation}\label{eq:GramConnection}
\text{for all $Q  \succeq 0$, $\| Q \|_{\operatorname{tr}} = \| X^T \! X \|_{\operatorname{tr}} = \operatorname{tr}( X^T\!X)
 = \| X \|_{F}^2$.}
\end{equation}
The key property of this norm for our purposes follows from the linear-algebraic notion of symmetric gauge functions (see, e.g.,~Horn \&~Johnson~(1990)).
\begin{lemma}[Dominance of Trace Norm]\label{prop:domTrNorm}
Amongst all unitarily invariant norms $\| \cdot \|$, we have that $\| \cdot \|_{\operatorname{tr}} \geq \| \cdot \|$.
\end{lemma}
Adopting this norm for problem~\ref{prob:kernel} therefore allows us to provide minimax arguments, and its unitary invariance implies the natural property that results depend only on the spectrum of the kernel $Q \succeq 0$ under consideration, just as in the case of the optimal rank-$k$ approximant $Q_k$.

To this end, note that any Schur complement is itself positive semi-definite.  Recalling from definition~\ref{defn:Nyst} that the error incurred by the Nystr\"om approximation is the norm of the corresponding Schur complement, and applying the definition of the trace norm as per~\eqref{eq:TraceNormEquivalence}, we obtain the following characterization of problem~\ref{prob:kernel} under trace norm.
\begin{proposition}[Nystr\"om Error in Trace Norm] \label{prop:NystErr}
Fix a subset $J \subset \lbrace 1,2,\ldots n \rbrace$ of cardinality $k < n$, and denote by $\bar J$ its complement in $\lbrace 1,2,\ldots n \rbrace$.  Then the error in trace norm induced by the Nystr\"om approximation of an $n$-dimensional kernel $Q\succeq 0$ according to definition~\ref{defn:Nyst}, conditioned on the choice of subset $J$, may be expressed as follows:
\begin{equation}\label{eq:NystErrTrace}
\| Q - \widetilde Q \|_{\operatorname{tr}} = \operatorname{tr}(Q_{\bar J \times \bar J}) - \operatorname{tr}(Q_{J \times \bar J}^T \, Q_{J \times J}^{-1}Q_{J \times \bar J})
\text{,}
\end{equation}
where ${J \times \bar J}$ denotes rows indexed by $J$ and columns by $\bar J$.
\end{proposition}

\begin{proof}
For any selected subset $J$ we have that the Nystr\"om error term is given by
\begin{equation*}
\| Q - \widetilde Q \| = \| Q_{\bar J \times \bar J} - Q_{J \times \bar J}^T \, Q_{J \times J}^{-1}Q_{J \times \bar J} \|
\end{equation*}
according to the notation of proposition~\ref{prop:NystErr}.  Now, the Schur complement of a positive semi-definite matrix is always itself positive semi-definite (see, e.g.,~Horn \&~Johnson~(1990)), and so the specialization of the trace norm for positive semi-definite norms, as per~\eqref{eq:TraceNormEquivalence}, applies.  We therefore conclude that
\begin{align}
\| Q_{\bar J \times \bar J} - Q_{J \times \bar J}^T \, Q_{J \times J}^{-1}Q_{J \times \bar J} \|_{\operatorname{tr}}
 & = \operatorname{tr}(Q_{\bar J \times \bar J} - Q_{J \times \bar J}^T \, Q_{J \times J}^{-1} Q_{J \times \bar J}) \label{eq:trnormerr}
\\ & = \operatorname{tr}(Q_{\bar J \times \bar J}) - \operatorname{tr}(Q_{J \times \bar J}^T \, Q_{J \times J}^{-1} Q_{J \times \bar J}) \notag
\text{.}
\end{align}
\end{proof}

While each term in the expression of proposition~\ref{prop:NystErr} depends on the selected subset $J$, if all elements of the diagonal of $Q$ are equal, then the term $\operatorname{tr}(Q_{\bar J \times \bar J})$ is constant.  This has motivated approaches to problem~\ref{prob:kernel} based on minimizing exclusively the latter term (Smola \&~Sch{\"o}lkopf~2000; Zhang \&~Kwok~2009).

We conclude with an illuminating proposition that follows from the Gram decomposition of~\eqref{eq:GramConnection}.
\begin{proposition}[Trace Norm as Regression Residual]\label{prop:regressResid}
Let $Q \succeq 0$ have the Gram decomposition $Q = X^T \! X$, and let $X$ be partitioned as $[ X_{J} \,\, X_{\bar J}]$ in accordance with proposition~\ref{prop:NystErr}.  Then the Nystr\"om error in trace norm of~\eqref{eq:NystErrTrace} is the error sum-of-squares obtained by projecting columns of $X_{\bar J}$ on to the closed linear span of columns of $X_J$.
\end{proposition}
\begin{proof}
If $Q$ is positive semi-definite, it admits the Gram decomposition $Q = X^T \! X$.  If we partition $X$ (without loss of generality) into selected and unselected columns $[ X_{J} \,\, X_{\bar J}]$ according to a chosen subset $J$, it follows that
\begin{equation*}
Q = X^T \! X = \begin{bmatrix} X_J^T X_J & X_J^T X_{\bar J} \\ (X_J^T X_{\bar J})^T &  X_{\bar J}^T X_{\bar J} \end{bmatrix} \text{.}
\end{equation*}
Therefore the $i$th diagonal of the residual error follows as
\begin{align*}
(Q_{\bar J \times \bar J} - Q_{J \times \bar J}^T \, Q_{J \times J}^{-1} Q_{J \times \bar J})_{ii}
 & = (X_{\bar J} ^T X_{\bar J} - X_{\bar J}^T X_J (X_J^T X_J)^{-1} X_J^T X_{\bar J}
)_{ii}
\\ & =  (X_{\bar J} ^T \left[I - X_J (X_J^T X_J)^{-1} X_J^T \right] X_{\bar J})_{ii}
\text{,}
\end{align*}
and hence the Nystr\"om error in trace norm is given by the sum of squared residuals obtained by projecting columns of $X_{\bar J} $ on to the space spanned by columns of $X_J$.
\end{proof}

\subsection{Annealed determinantal distributions}

With this error characterization in hand, we may now define and introduce the notion of \emph{annealed determinantal distributions}, which in turn provides a framework for the analysis and comparison of landmark selection and sampling methods.
\begin{defn}[Annealed Determinantal Distributions]\label{def:detAnneal}
Let $Q \succeq 0$ be a positive semi-definite kernel of dimension $n$, and fix an exponent $s \geq 0$.  Then, for fixed $k\leq n$, $Q$ admits a family of probability distributions defined on the set of all $J \subset \lbrace 1,2,\ldots n \rbrace: |J|=k$ as follows:
\begin{equation}\label{eq:anneal}
p^s(J) \propto \det(Q_{J \times J})^s; \quad s \geq 0, \,\, |J| = k \text{.}
\end{equation}
\end{defn}
This distribution is well defined because all principal submatrices of a positive semi-definite matrix are themselves positive semi-definite, and hence have nonnegative determinant.  The term \emph{annealing} is suggestive of its use in stochastic computation and search, where a probability distribution or energy function is gradually raised to some nonnegative power over the course of an iterative sampling or optimization procedure.

Indeed, for $0<s<1$ the determinantal annealing of definition~\ref{def:detAnneal} amounts to a flattening of the distribution of $\det(Q_{J \times J})$, whereas for $1<s<\infty$ it becomes more and more peaked.  In the limiting cases we recover, of course, the uniform distribution on the range of $\det(Q_{J \times J})$, and respectively mass concentrated on its maximal element(s).

It is instructive to consider these limiting cases in more detail.  Taking $s=0$, we observe that the method of uniform sampling typically favoured by practitioners (Smola \&~Sch{\"o}lkopf~2000; Fowlkes {\em et al.}~2001, 2004; Talwalkar {\em et al.}~2008) is trivially recovered, with negligible associated computational cost.  By extending the result of Belabbas \&~Wolfe~(2009), the induced error may be bounded as follows.
\begin{theorem}[Uniform Sampling]\label{prop:unifSamp}
Let $Q \succeq 0$ have the Nystr\"om extension $\widetilde Q$, where subset $J:|J|=k$ is chosen uniformly at random.  Averaging over this choice, we have
\begin{equation*}
\mathbb{E} \|Q - \widetilde Q \|_{\operatorname{tr}} \leq \frac{n-k}{n} \operatorname{tr}(Q)
\text{.}
\end{equation*}
\end{theorem}
Note that this bound is tight, with equality attained for diagonal $Q \succeq 0$.  Uniform sampling thus averages the effects of \emph{all} eigenvalues of $Q$, in contrast to the optimal rank-$k$ approximation obtained by retaining the $k$ principal eigenvalues and eigenvectors from an exact spectral decomposition, which incurs an error in trace norm of only $\sum_{i = k + 1}^n \lambda_i$.

In contrast to annealed determinant sampling, uniform sampling fails to place zero probability of selection on subsets $J$ such that $\det(Q_{J \times J}) = 0$.  As the following proposition of Belabbas \&~Wolfe~(2009) shows, the exact reconstruction of rank-$k$ kernels from $k$-subsets via the Nystr\"om completion requires the avoidance of such subsets.
\begin{proposition}[Perfect Reconstruction via Nystr\"om Extension]\label{prop:perfectReconst}
Let $Q \succeq 0$ be $n \times n$ and of rank $k$, and suppose that a subset $J:|J|=k$ is sampled according to the annealed determinantal distribution of definition~\ref{def:detAnneal}.  Then, for all $s > 0$, the error $\|Q - \widetilde Q \|_{\operatorname{tr}}$ incurred by the Nystr\"om completion of~\eqref{eq:partQtilde} will be equal to zero.
\end{proposition}
\begin{proof}
Whenever $\operatorname{rank}(Q) = k$, only full-rank (i.e., rank-$k$) principal submatrices of $Q$ will be nonsingular, and hence admit nonzero determinant.  Therefore, for any $s>0$, these will be the only submatrices selected by the annealed determinantal sampling scheme.  By proposition~\ref{prop:regressResid}, the full-rank property implies that the regression error sum-of-squares will in this case be zero, implying that $ \widetilde Q = Q$.
\end{proof}

Considering the limiting case as $s \to \infty$, we equivalently recover the problem of \emph{maximizing} the determinant, which is well known to be $\mathit{NP}$-hard.  Since
$\operatorname{det}(Q) = \operatorname{det} (Q_{J \times J}) \times \operatorname{det} (Q_{\bar J \times \bar J} - Q_{J \times \bar J}^T \, Q_{J \times J}^{-1}Q_{J \times \bar J})$, the notion of subset selection based on maximal determinant admits the following interesting correspondence, since, if $x$ is a vector-valued Gaussian random variable with covariance matrix $Q$, then the Schur complement of $Q_{J \times J}$ in $Q$ is the conditional covariance matrix of components $x_{\bar J}$ given $x_{J}$.
\begin{proposition}[Minimax Relative Entropy]\label{prop:relEnt}
Fix an $n$-dimensional kernel $Q \succeq 0$ as the covariance matrix of a random vector $x\in\mathbb{R}^n$ and fix an integer $k<n$.  Minimizing the maximum relative entropy of coordinates $x_{\bar J}$, conditional upon having observed coordinates $x_{J}$, corresponds to selecting $J$ such that $\operatorname{det}(Q_{J \times J})$ is maximized.
\end{proposition}
\begin{proof}
The Schur complement $S_C(Q_{J \times J})$ represents the covariance matrix of $x_{\bar J}$ conditional upon having observed $x_{J}$.  To this end, we first note the following relationship (Horn \&~Johnson~1990):
\begin{equation*}
\operatorname{det}(Q) = \operatorname{det} (Q_{J \times J}) \times \operatorname{det} (S_C(Q_{J \times J})) \text{.}
\end{equation*}
Moreover, for fixed covariance matrix $Q$, the multivariate Normal distribution maximizes entropy $h(x)$, and hence for $S_C(Q_{J \times J})$ we have that
\begin{equation*}
h(x_{\bar J}\,\vert\,x_{J}) = \frac{1}{2}\log \frac{|2\pi e \,Q |}{|2\pi e \,Q_{J \times J} |} =  \log |2\pi e \,S_C(Q_{J \times J}) |^{\frac{1}{2}}
\end{equation*}
is the maximal relative entropy attainable upon having observed $x_{J}$.
\end{proof}
To this end the bound of Goreinov \&~Tyrtyshnikov~(2001) extends to the case of the trace norm as follows.
\begin{theorem}[Determinantal Maximization]\label{prop:DetMaxim}
Let $\widetilde{Q}$ denote the Nystr\"om completion of a kernel $Q \succeq 0$ via subset $J = \operatornamewithlimits{argmax}_{J':|J'|=k} \operatorname{det}(Q_{J' \times J'})$.  Then
\begin{equation*}
 \|Q-\widetilde{Q} \|_{\operatorname{tr}} \leq (k+1) \, (n-k) \lambda_{k+1}(Q)
 \text{,}
\end{equation*}
where $\lambda_{k+1}(Q)$ is the $(k+1)$th largest eigenvalue of $Q$.
\end{theorem}

We conclude with a recent result (Belabbas \&~Wolfe~2009) bounding the expected error for the case $s=1$, that in turn improves upon the additive error bound of Drineas \&~Mahoney~(2005) for sampling according to the squared diagonal elements of $Q$.
\begin{theorem}[Determinantal Sampling]\label{prop:volSamp}
Let $Q \succeq 0$ have the Nystr\"om extension $\widetilde Q$, where subset $J:|J|=k$ is chosen according to the annealed determinantal distribution of~\eqref{eq:anneal} with $s=1$.  Then
\begin{equation*}
\mathbb{E} \|Q - \widetilde Q \|_{\operatorname{tr}} \leq (k+1)
\sum_{i=k+1}^n\lambda_i(Q) \text{,}
\end{equation*}
with $\lambda_{i}(Q)$ the $i$th largest eigenvalue of $Q$.
\end{theorem}
This result can be related to that of theorem~\ref{prop:DetMaxim}, which depends on  $n-k$ times $\lambda_{k+1}(Q)$, the $(k+1)$th largest eigenvalue of $Q$, rather than the sum of its $n-k$ smallest eigenvalues.  It can also be interpreted in terms of the \emph{volume sampling} approach proposed by Deshpande {\em et al.}~(2006), applied to the Gram matrix $X_J^TX_J$ of an `arbitrary' matrix $X_J$, as $\det(Q_{J \times J}) = \det(X_J^TX_J) = \det(X_J)^2$.  By this same argument, Deshpande {\em et al.}~(2006) show the result of theorem~\ref{prop:volSamp} to be  essentially the best possible.

We conclude by noting that, for most values of $s$, sampling from the distribution $p^s(J)$ presents a combinatorial problem, because of the $\binom{n}{k}$ distinct $k$-subsets associated with an $n$-dimensional kernel $Q$.  To this end, a simple Markov chain Monte Carlo method has been proposed by Belabbas \&~Wolfe~(2009) and shown to be effective for sampling according to the determinantal distribution on $k$-subsets induced by $Q$.  This Metropolis algorithm can easily be extended to the cases covered by definition~\ref{def:detAnneal} for all $s \geq 0$. We also note that tridiagonal approximations to $\det(Q_{J \times J})$ can be computed in $\mathcal{O}(k)$ operations, and hence offer an alternative to the $\mathcal{O}(k^3)$ cost of exact determinant computation.

\section{Case study: application to computer vision}
\label{sec:caseStudy}

In the light of the range of methods described above for optimizing the landmark selection process through sampling, we now consider a case study drawn from the field of computer vision, in which low-dimensional manifold structure is extracted from high-dimensional video data streams.  This field provides a particularly compelling example, as algorithmic aspects, both of space and time complexity, have historically had a high impact on the efficacy of computer vision solutions.

Applications in areas as diverse as image segmentation (Fowlkes {\em et al.}~2004), image matting (Levin {\em et al.}~2008), spectral mesh processing (Liu {\em et al.}~2006) and object recognition through the use of appearance manifolds (Lee \&~Kriegman~2005) all rely in turn on the eigendecomposition of a suitably defined kernel.  However, at a complexity of $\mathcal{O}(n^3)$ the full spectral analysis of real-world datasets is often prohibitively costly---requiring in practice an approximation to the exact spectral decomposition.  Indeed the aforementioned tasks typically fall into this category, and several share the common feature that their kernel approximations are obtained in exactly the same way---via the process of selecting a subset of landmarks to serve as a basis for computation.

\subsection{The spectral analysis of large video datasets}

\begin{figure}[t]
\centering
\includegraphics[width=.9\columnwidth]{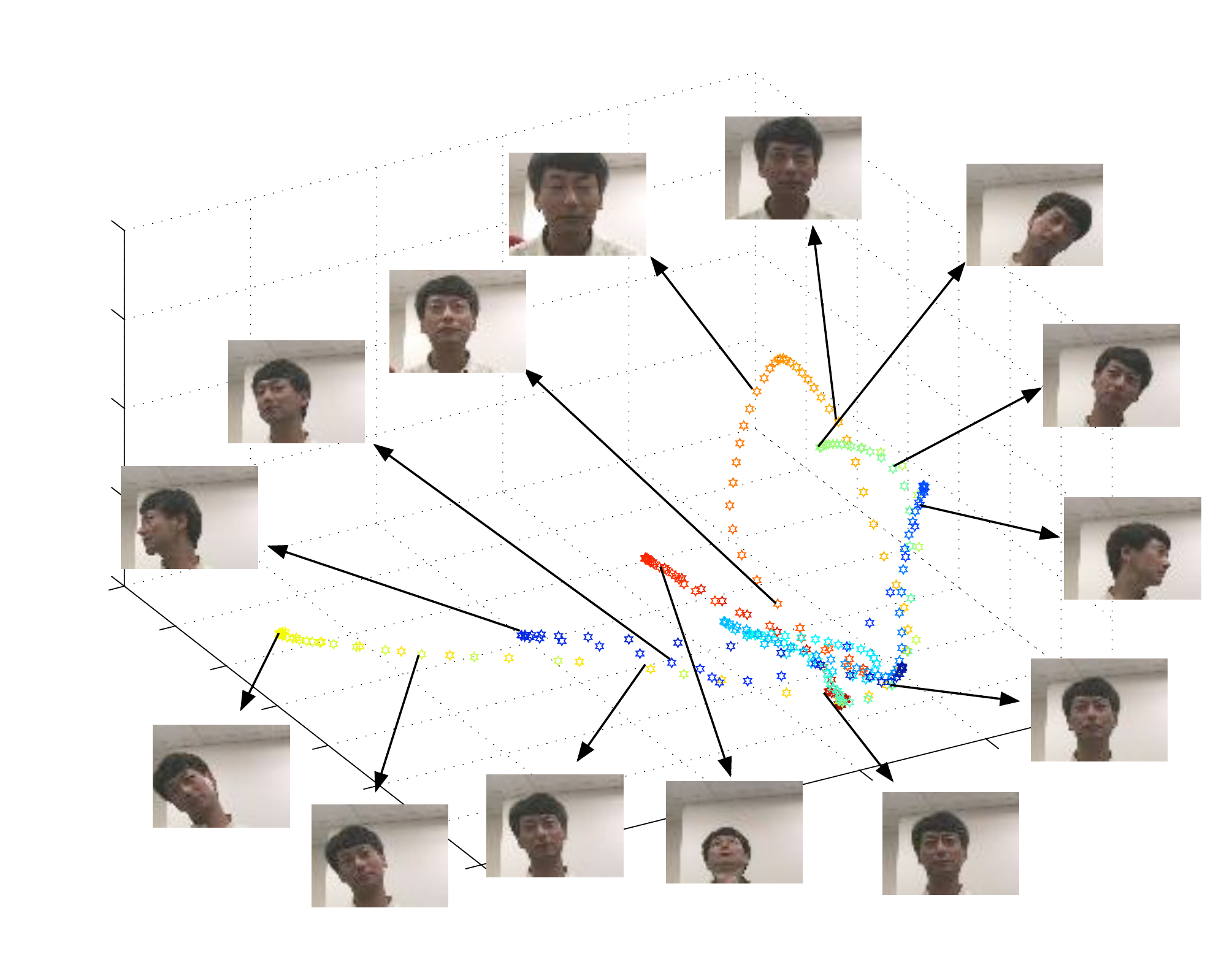}
  \caption{\label{fig_fuji_embed}Diffusion maps embedding of the video \texttt{fuji.avi} from the Honda/UCSD
  database (Lee {\em et al.}~2003, 2005), implemented in the pixel domain after data normalization,
  $\sigma=100$.}
\end{figure}

Video datasets may often be assumed to have been generated by a
dynamical process evolving on a low-dimensional manifold, for example  a line in the case of a
translation, or a circle in the case of a  rotation. Extracting this low-dimensional space has
applications in object recognition through appearance manifolds (Lee {\em et~al.}, 2005), motion
recognition (Blackburn \&~Ribeiro~2007), pose estimation (Elgammal \&~Lee~2004) and others.  In this context, nonlinear dimension reduction algorithms (Lin \&~Zha~2008) are the
key ingredient mapping the video stream to a lower-dimensional space. The vast majority of
these algorithms require one to obtain the eigenvectors of a positive definite kernel $Q$
of size equal to the number of frames in the video stream, which quickly becomes
prohibitive and entails the use of approximations to the exact spectral analysis of $Q$.

To begin our case study, we first tested the efficacy of the Nystr\"om extension coupled with the subset selection
procedures given in~\S\ref{sec:analysisFramework} on different video datasets. In
figure~\ref{fig_fuji_embed} we show the exact embedding in three dimensions, using the diffusion maps
algorithm (Coifman {\em et al.}~2005),  of a video from the Honda/UCSD database (Lee {\em et al.}~2003, 2005), as well as some selected frames. In  this video,
the subject rotates his head in front of the camera in several directions, with each motion
starting from the resting position of looking straight at the camera. We observe that with
each of these motions is associated a circular path, and that they all originate from   the
same area (the lower-front-right area) of the graph, which corresponds to the resting position.

\begin{figure}[t] \centering
 \includegraphics[width=0.67\columnwidth]{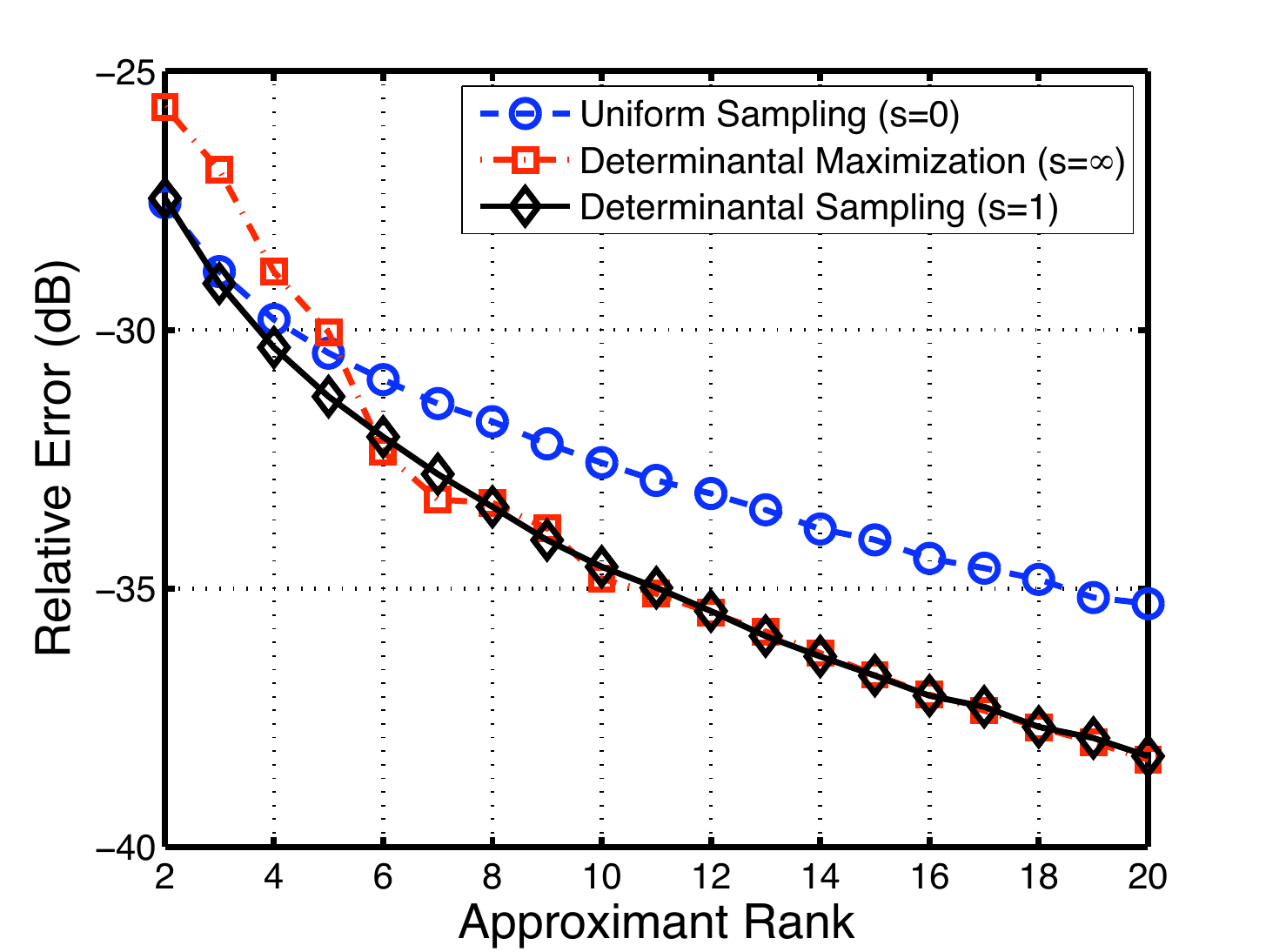}
  \caption{\label{fig_error_vid_fuji}Average normalized approximation error of the
  Nystr\"om reconstruction of the diffusion maps kernel obtained from the video of
  figure~\ref{fig_fuji_embed} using different subset selection methods. Sampling according
  to the determinant yields overall the best performance.}
  \end{figure}

In figure~\ref{fig_error_vid_fuji}, the average approximation error for the
diffusion maps kernel corresponding to this video is evaluated, for an approximation rank between 2 and
20. The results are averaged over 2000 trials. The sampling from the determinant
distribution is done via a Monte Carlo algorithm similar
to that of Belabbas \&~Wolfe~(2009) and the determinant maximization is
obtained by keeping the subset $J$ with the largest corresponding determinant $Q_{J}$ over
a random choice of 2500 subsets. For this setting, sampling according to the determinant
distribution yields the best results uniformly across the range of approximations. We
observe that keeping the subset with maximal determinant does not give a good approximation
at low ranks. A further analysis showed that in this case the chosen landmarks tend to
concentrate around the lower-front-right region of the graph, which yields a good
 approximation locally in this part of the space but fails to recover other regions properly. This behaviour illustrates
the appeal of randomized methods, which avoid such pitfalls.

\begin{figure}[t]
\centering
\includegraphics[width=\columnwidth]{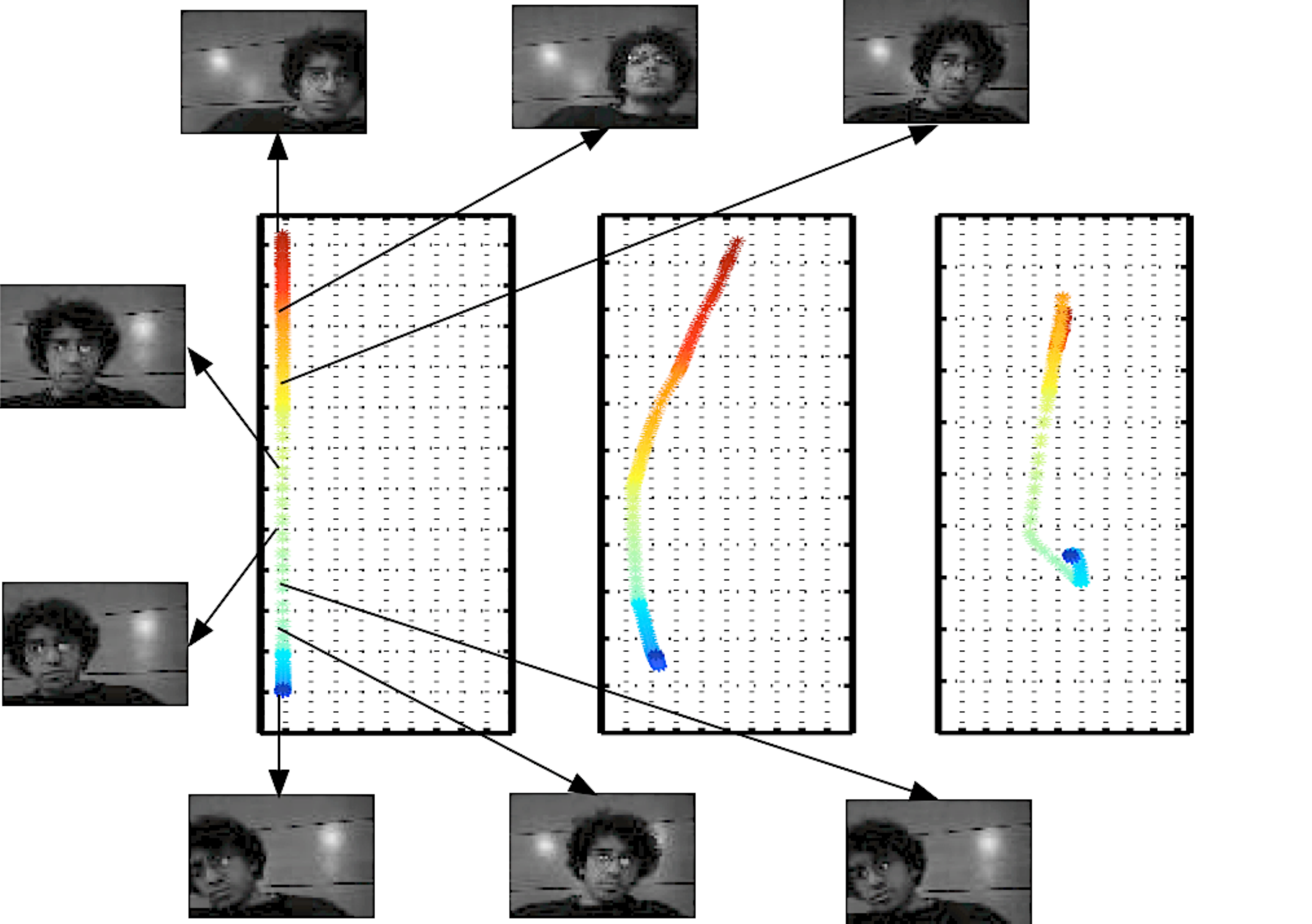}
  \caption{\label{fig_ali_embed}Exact (left), determinant sampling (centre), uniform sampling (right) diffusion maps embedding of a video showing the movement at an uneven speed, implemented in the pixel domain with $\sigma=100$ after data normalization.  Note that the linear structure of this manifold is recovered almost exactly by the determinantal sampling scheme, whereas it is lost in the case of uniform sampling, where the curve folds over on to itself.}
\end{figure}

As a subsequent demonstration, we collected video data of the first author moving slowly in front of a camera at an uneven speed.  The resulting embedding, given again by the diffusion maps algorithm, is a non-uniformly
sampled straight line. In this case, we can thus evaluate by visual inspection the effect
of an approximation of the diffusion map kernel on the quality of the embedding. This is
shown in figure~\ref{fig_ali_embed}, where  typical results from  different subset selection
methods are displayed. We see that sampling according to the determinant recovers the
linear structure of the dynamical process, up to an affine transformation, whereas sampling
uniformly yields some folding of the curve over itself at the extremities and  centre.

\begin{figure}[t]  \centering \includegraphics[width=0.67\columnwidth]{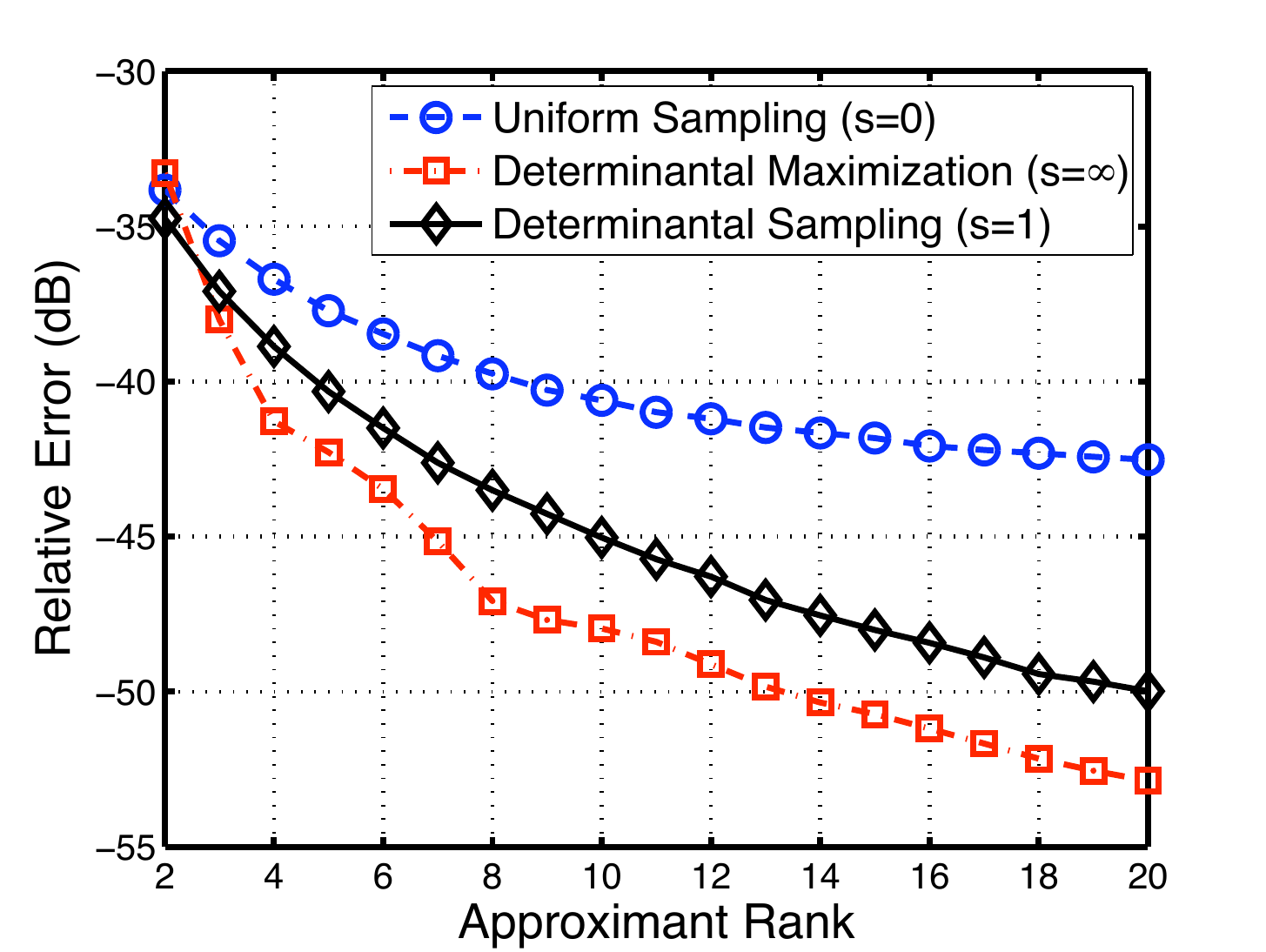}
  \caption{\label{fig_error_vid_ali}Average normalized approximation error of the Nystr\"om
  reconstruction of the diffusion maps kernel obtained from the video of
  figure~\ref{fig_ali_embed} using different subset selection methods. Sampling uniformly
  is consistently outperformed by the other methods.}
\end{figure}

In figure~\ref{fig_error_vid_ali}, we show the approximation error of the kernel associated
with this video averaged over 2000 trials, similarly to the previous example. In this case,
maximizing the determinant yields the best overall performance. We observe that sampling
according to the determinant easily outperforms choosing the subset uniformly at random, lending further credence to our analysis framework and its practical implications for landmark selection and data subsampling.

\begin{acknowledgements}

This material is based in part upon work supported by the Defense Advanced Research Projects Agency under  Grant~HR0011-07-1-0007, by the National Institutes of Health under Grant~P01~CA134294-01, and by the National Science Foundation under Grants~DMS-0631636 and~\mbox{CBET-0730389}.  Work was performed in part while the authors were visiting the Isaac Newton Institute for Mathematical Sciences under the auspices of its programme on statistical theory and methods for complex high-dimensional data, for which support is gratefully acknowledged.

\end{acknowledgements}

\label{lastpage}
\end{document}